# Extending the adverbial coverage of a NLP oriented resource for French

**Elsa Tolone**
LIGM, Université Paris-Est, France
FaMAF, Universidad Nacional de Córdoba, Argentina
`tolone@univ-paris-est.fr`

**Stavroula Voyatzi**
LIGM, Université Paris-Est, France
`voyatzi@univ-mlv.fr`

## Abstract

This paper presents work on extending the adverbial entries of *LGLex*, a NLP oriented syntactic resource for French. Adverbs were extracted from the Lexicon-Grammar tables of both simple adverbs ending in *–ment* '–ly' (Molinier and Levrier, 2000) and compound adverbs (Gross, 1986; 1990). This work relies on the exploitation of fine-grained linguistic information provided in existing resources. Various features are encoded in both LG tables and they haven't been exploited yet. They describe the relations of deleting, permuting, intensifying and paraphrasing that associate, on the one hand, the simple and compound adverbs and, on the other hand, different types of compound adverbs. The resulting syntactic resource is manually evaluated and freely available under the LGPL-LR license.

## 1 Introduction

Recognising adverbs such as *extrêmement* 'extremely' and *à long terme* 'in the long run' in texts is likely to be useful for information retrieval and extraction because of the information that some of these adverbials convey.

Adverbs, or more generally circumstantial complements, have often been overlooked in the compilation of lexical resources (Nølke, 1990: 3). Several reasons explain this lack of interest. Firstly, adverbials are usually felt as less useful than nouns for information retrieval and extraction. Secondly, compound adverbs in particular are difficult to distinguish from prepositional phrases assuming other syntactic functions, such as arguments or noun modifiers: the distinction is hardly correlated to any material markers in texts and lies in complex linguistic notions (Villavicencio, 2002; Merlo, 2003).

The availability of large-coverage lexicons providing lexical, syntactic and semantic information is essential in order to gain insight on the recognition of adverbs, including the dual problems of variability and ambiguity. In addition, it is likely to help solving prepositional phrase attachment during shallow or deep parsing (Agirre *et al.*, 2008).

In this paper, we present work on extending the adverbial entries of *LGLex*, a NLP oriented syntactic resource for French. These adverbs were extracted from the Lexicon-Grammar tables (hereafter LG tables) of both simple adverbs ending in *–ment* '–ly' (Molinier and Levrier, 2000) and compound adverbs (Gross, 1986; 1990).

The paper is organized as follows. In section 2, we provide an overview of the three resources used in our work. In section 3, we describe how we enhanced *LGLex* thanks to various features that are encoded in LG tables. In section 4, we present results and discuss the manual evaluation process. Finally, we conclude in section 5 by pointing out several possible extensions and issues for future research.

## 2 Resources

Lexicon-Grammar tables (hereafter LG tables) are currently one of the major sources of lexical syntactic information for the French language; several LG tables exist for other languages (see section 2.1). Their development was initiated as early as the 1970s by Maurice Gross, at the LADL (Gross, 1975; 1994), and then at the LIGM, University Paris-Est (Boons *et al.*, 1976; Guillet and Leclère, 1992).

Lexical information is represented in the form of tables. Each table puts together elements of a given lexical-grammatical category (for a given language) that share a certain number of defining features, which usually concern subcategorization information. These elements form a class. These tables are represented as matrices (see section 2.1.2): each row corresponds to a lexical item of the corresponding class; each column lists all features that may be valid or not for the different members of the class; at the intersection of a row and a column, the symbol + (resp. –) indicates that the feature corresponding to the column is valid (resp. not valid) for the lexical entry corresponding to the row.

The resources described in this paper correspond to the LG tables of both simple and compound adverbs, in which previously implicit features have been made explicit[1] for more convenient use in NLP. All tables are fully available[2] under a free license (LGPL-LR).

### 2.1 The LG tables of adverbs

LG tables of adverbs are available in several languages, notably in English (Gross, 1986), German (Seelbach, 1990), Spanish (Blanco and Català, 1998/1999), Italian (De Gioia, 2001), Portuguese (Baptista, 2003), Korean (Jung, 2005) and Modern Greek (Voyatzi, 2006).

In French, there are two resources of adverbs that follow different principles both in classification and in representation within the Lexicon-Grammar framework. That is, firstly, tables of simple adverbs ending in *–ment* '–ly' (Moliner, 1984; Molinier and Lévrier, 2000), which are mainly derived from adjectives and, secondly, tables of compound adverbs (Gross, 1986; 1990). In this section, we describe briefly the different classes, morphosyntactic structures and features provided in the tables of both types of adverbs.

#### 2.1.1 The LG tables of simple adverbs

According to Molinier and Levrier (2000), adverbs ending in *–ment* '–ly' form a large class of French adverbs. Moreover, as opposed to other types of adverbs, they form a quite open class.

These adverbs constitute a morphologically homogeneous class, since most of them are created according to the pattern `adjective + -ment` '-ly'. In total, there are 3,203 simple adverbial entries which are represented in sixteen LG tables.

The first partition has been established between sentential adverbs and adverbs integrated into the sentence; that means attached to the predicate or any other component of the sentence.

Three major classes of sentential adverbs are worth mentioning: conjuncts, style disjuncts (or utterance-level adverbs) and attitude disjuncts (or statement adverbs). The latter are divided into four classes: evaluative adverbs, e.g. *curieusement* 'curiously', adverbs of habit, e.g. *habituellement* 'usually', modal adverbs, e.g. *certainement* 'certainly', and subject oriented attitude adverbs, e.g. *bêtement* 'foolishly'.

On the other hand, there are six major classes of adverbs integrated into the sentence:
(i) adverbs of subject oriented manner, e.g. *gentiment* 'kindly', *Max traite les gens gentiment = Max est gentil dans la (façon + manière) dont il traite les gens* 'Max treats people kindly = Max is kind in the way he treats people';
(ii) adverbs of verbal manner, e.g. *démocratiquement* 'democratically', *Ce parti est arrivé au pouvoir démocratiquement = C'est démocratiquement que ce parti est arrivé au pouvoir* 'This party came to power democratically = It is democratically that this party came to power';
(iii) adverbs of quantity (including intensifiers), e.g. *excessivement* 'excessively';
(iv) adverbs of time expressing time, e.g. *actuellement* 'actually', duration, e.g. *temporairement* 'temporarily' or frequency, e.g. *régulièrement* 'regularly';
(v) viewpoint adverbs, e.g. *linguistiquement = d'un point de vue linguistique* 'linguistically = from a linguistic point of view';
(vi) focus adverbs, e.g. *essentiellement* 'essentially'.

Features included in the LG tables can be organized into four main groups: distributional features (e.g. the possibility for an adverb to occur at the beginning of a negative clause), local syntactic features (e.g. the possibility for an adverb to have a function as an indefinite determiner), semantic features (e.g. knowing what type of interrogative adverb corresponds to an adverb allows it to be categorized semantically), and paraphrasing features (e.g. *Adv, Pindicatif = C'est Adj que Psubjonctif =: Bizarrement, Marie n'est pas venue à la soirée = C'est bizarre que Marie ne soit pas venue à la soirée* ' Oddly, Mary did not come to the party = It is odd that Mary did not come to the party').

#### 2.1.2 The LG tables of compound adverbs

The scope of the LG tables of compound adverbs is delimited by the intersection of two criteria: (i) multiword expressions and (ii) adverbial func-tion. As Laporte and Voyatzi (2008) state "a phrase composed of several words is considered to be a multiword expression if some or all of its components are tied together, that is, if their combination does not obey productive rules of syntactic and semantic compositionality". This criterion ensures a complementarity between lexicon and grammar. In other words, it tends to ensure that any combination of linguistic elements which is correct in the language, but is not represented in common syntactic-semantic grammars, should be stored in lexicons.

---
[1] In order to make previous implicit features explicit, a table of classes has been created (Tolone, 2009; 2011). Its role is to assign features when their value is constant over a class, e.g. class definition features. Each row stands for a class and each column stands for a feature. Each cell corresponds to the validity of a feature in a class. In particular, the table of French adverb classes is composed of 32 different classes and 163 features.
[2] http://infolingu.univ-mlv.fr/english > Language Resources > Lexicon-Grammar > Download.

In terms of the adverbial function, LG tables of compound adverbs deal only with circumstantial complements, namely, complements that modify the predicate or any other element of the sentence in which they occur. Sentential adverbs modify or enhance the entire sentence. They are identified through criteria (Gross, 1986; 1990) involving the fact that they are optional, they combine freely with a wide range of predicates and some of them typically answer questions such as *comment?* 'how?', *où?* 'where?', *quand?* 'when?', etc. The compound adverbs described in LG tables (Gross, 1986; 1990) take several morphosyntactic forms: unsuffixed adverbs, e.g. *demain*[3] 'tomorrow', suffixed adverbs, e.g. *nuitamment*[4] 'by night', prepositional phrases, e.g. *à la dernière minute* 'at the last minute', noun phrases, e.g. *le cas échéant* 'if necessary' or adverbial clauses, e.g. *jusqu'à ce que mort s'ensuive* 'until death comes'.

These compound adverbs are classified according to their internal morphosyntactic structure which is described at the elementary level of sequences of parts of speech. The classification is based mainly on the number, type and position of the fixed and variable lexical components of adverbs. These classificatory morphosyntactic structures have a three-fold aim: first, they are intended as an aid to organize the heterogeneous compound adverbs in an electronic lexicon; second, they are intended as an aid to identify the compound adverbs in a parser; finally, they have impact on the syntactic-semantic subcategorization of compound adverbs. For instance, most of the adverbial clauses represented in table PF are interpolated clauses, e.g. *si ma mémoire est bonne* 'if my memory serves me well'.

Table 1 illustrates the sixteen formal classes, together with their defining internal morphosyntactic structure, an illustrative example[5] and the number of entries listed:

| Identifier | Morphosyntactic structure | Example | Size |
|---|---|---|---|
| PADV | C | enfin | 520 |
| PC | Prép C | par exemple | 664 |
| PDETC | Prép Det C | de nos jours | 848 |
| PAC | Prép Det Adj C | à la dernière minute | 776 |
| PCA | Prép Det C Modif pré-adj Adj | à la nuit tombante | 840 |
| PCDC | Prép Det1 C1 de Det2 C2 | dans la limite du possible | 750 |
| PCPC | Prép Det1 C1 Prép2 Det2 C2 | à cent pour cent | 287 |
| PCONJ | Prép Det1 C1 Conj Prép2 Det2 C2 | tôt ou tard | 333 |
| PCDN | Prép Det1 C1 de N2 | à l'insu de N | 555 |
| PCPN | Prép Det1 C1 Prép2 N2 | en comparaison avec N | 151 |
| PV | Prép V Prépv Detv Cv | à dire vrai | 285 |
| PF | ConjS (Det0 C0 + N0) V Prép1 (Det1 C1 + N1) | jusqu'à ce que mort s'ensuive | 396 |
| PECO | (Adj) comme Det C | (fidèle) comme un chien | 305 |
| PVCO | (V) comme Det C | (travailler) comme un chien | 338 |
| PPCO | comme Prép Det C | (disparaître) comme par enchantement | 50 |
| PJC | Conjc Det C1 Prép C2 | mais aussi et surtout | 185 |
| | | Total | 7,283 |

Table 1. Morphosyntactic structures of French compound adverbs

Compound adverbs are represented in sixteen LG tables[6], one for each of the defining morphosyntactic structures[7]. Unlike simple adverbs, compound adverbs are represented in the tables within a structure of elementary sentence which is composed of a verbal predicate (intransitive in most cases) and its arguments. This representation takes into account the combination of the

---

[3] According to Gross (1990: 153), *demain* 'tomorrow' is a compound form from an etymological point of view: from the Latin expression *de mane* which means literally "in the morning". But, it is now regarded as a simple form since it is represented by a single word. These forms are derived from noun or prepositional noun phrases previously analyzed that were tied at various times.

[4] According to Gross (1986: 2), *nuitamment* 'by night' can also be considered as an idiomatic compound, though not constituted of words but of a word and a suffix. Lack of compositionality stems from the observation that *quotidiennement* 'daily', *mensuellement* 'monthly', etc. which are derived adverbs of the same formal type have a regular formation, in the sense that their interpretation is homogeneous. Thus, *nuitamment* 'by night' is an isolated case, as opposed to an open series of identical forms with a different interpretation.

[5] It is possible that one or more of the components defined in a morphosyntactic structure are absent. For instance, in the adverb *à cent pour cent* 'one hundred percent', which is assigned the structure PCPC, Dét1 and Dét2 are empty.

[6] The LG tables of adverbs described in this paper have been updated by the members of LIGM, so that previously implicit linguistic information becomes explicit and convenient for NLP applications. Tolone (2009; 2011) and Tolone *et al*. (2010) give a thorough account of this work.

[7] Symbols with obvious interpretation are used such as: `Prép` (preposition), `Det` (determiner), `Adj` (adjective), `Modif pré-adj` (pre-adjectival modifier), `N` (noun), `V` (verb), `Conj` (conjunction), and `C` (noun tied with the rest of the adverbial).

adverb with a structure of elementary sentence, and, thus provides precise information about various types of constraints occurring between compound adverbs and the predicate they modify. An example of time constraint is given below:

*Les tablettes (remplaceront + *ont remplacé + *remplacent) les PC dans un avenir proche*

'Ipads (will replace + *replaced + *replace) desktop PC in a near future'

The LG tables of French compound adverbs contain 7,283 entries. Table 2 displays a sample of the table PCA which is defined by the morpho-syntactic structure `Preposition, Determiner, Constrained noun, Pre-adjectival Modifier, Adjective`:

| N0 = Nhum | N0 = N-num | Neg obl | Ppv | Prédicat type | <ENT>Prép | <ENT>Det | <ENT>C | <ENT>Modif pré-adj | <ENT>Adj | Prép Det C | Prép Det Modif pré-adj Adj C | Prép Modif pré-adj Adj C | Conjonction |
|---|---|---|---|---|---|---|---|---|---|---|---|---|---|
| - | + | - | :se | produire | dans | le | cas | <E> | contraire | - | - | - | + |
| - | + | - | :se | produire | <E> | le | cas | <E> | échéant | - | - | - | - |
| - | + | - | :se | produire | à | le | cas | <E> | où | - | - | - | - |
| - | + | - | :se | produire | dans | le | cas | qui | préoccupe ":Nhum" | - | - | - | - |
| - | + | - | :se | produire | dans | le | cas | <E> | présent | - | + | - | + |
| - | + | - | :se | produire | dans | un | cas | <E> | semblable | - | + | - | - |

Table 2. Compound adverbs of table PCA

In this table, each row corresponds to a lexical item with adverbial function, and each column corresponds to:

- one of the components in the morphosyntactic structure of the items, i.e. features with identifiers `Prép, Det, C, Modif pré-adj,` and `Adj`;

- a syntactic feature holding binary values, for example: `Prép Det Modif pré-adj Adj C` describes the possible permutation (without loss of information) of the adjectival phrase represented in this table as `Modif pré-adj Adj`; moreover, `Neg obl` encodes the constraint that the adverbial occurs obligatorily in a negative clause;

- a semantic feature holding binary values, for instance, `Conjonction` points out whether the compound adverb has a connector function in discourse, i.e. it links the clause in which it occurs with the previous clause as, for example, *dans le cas contraire* 'otherwise';

- an item of information provided as an aid to help human readers find examples of sentences containing the compound adverb: features with identifiers `Ppv` and `Prédicat type` give an example of a verbal predicate that combines commonly with the adverb.

Unlike simple adverbs, the sixteen classes of compound adverbs, represented in sixteen LG tables, are both syntactically and semantically heterogeneous. For instance, the table PAC encodes adverbs that are defined by the morpho-syntactic structure `Prép Det Adj C`, but belong to different syntactic and semantic classes of Molinier and Lévrier (2000): conjuncts, e.g. *dans un premier temps* 'initially', disjuncts, e.g. *à Poss0 humble avis* 'in Poss0 humble opinion', adverbs of time, e.g. *depuis cent sept ans* 'one hundred and seven years since', adverbs of verbal manner, e.g. *n'importe comment* 'no matter how', etc.

Despite their differences, both types of adverbs are often related by productive and regular relations such as, for example, paraphrasing relations allowing the creation of pairs of synonyms, as shown in Table 3:

| Adverbs encoded in LG tables of simple adverbs | Adverbs encoded in LG tables of compound adverbs |
|---|---|
| *pratiquement* (ADVPS) 'practically' | *en pratique* (PC) 'in practice' |
| *franchement* (ADVPS) 'frankly' | *à franchement parler* (PV) 'frankly speaking' |
| *sincèrement* (ADVMS) 'sincerely' | *de (E+une) (manière+façon) sincère* (PCA) 'in a sincere way' |
| *politiquement* (ADVMP) 'politically' | *du point de vue politique* (PCA) 'from a political point of view' |

Table 3. Paraphrasing relations between simple and compound adverbs

### 2.2 The syntactic lexicon *LGLex*

The current version of the French LG tables has to consider the use of these lexical data in NLP tools (Tolone, 2009). Therefore, the tables have been converted into an interchange format, based on the same linguistic concepts as those handled in the tables. This conversion is based on *LGExtract*: a generic tool for generating a syntactic lexicon for NLP from the LG tables (Constant and Tolone, 2010). It relies, first, on a global table of classes in which we added the missing features and, second, on a single extraction script including all operations related to each feature to be performed for all tables.

Thanks to *LGExtract*, a French lexicon for NLP has been generated from all LG tables and for most lexical-grammatical categories: verbs, predicative nouns, idioms and adverbs. This syntactic lexicon is named *LGLex* (Constant and Tolone, 2010; Tolone, 2011) and it is freely available[8] under the LGPL-LR license in both plain text format and XML.

*LGLex* is currently composed of 13,867 verbal entries (from 67 tables), 12,696 nominal entries (from 78 tables), 39,628 idioms (from 69 tables) and 10,487 adverbial entries (from 32 tables) of which 3,203 are simple adverbs (from 16 tables) and 7,284 are compound adverbs (from 16 tables).

# 3 Extending the LGLex

Each entry of the lexicon includes three sections:

(i) section `Lexical information` identifies the lexical entry, for instance, the adverb *jusqu'à la fin des* (=de les) *temps* 'until the end of time' which is encoded in table PCDC, and also gives the category of each lexical component. We added the information of `paraphrases`, `other structures` and `other entries with intensification` (see section 3.2);

(ii) section `Arguments` gives information about the arguments of the predicate: for instance, the subject argument `N0`, assigned to the predicate that can be modified by the adverb *jusqu'à la fin des temps* 'until the end of time' is a non human noun phrase, represented by `N0 = N-hum`;

(iii) section `Constructions` enumerates the identifiers of all constructions of the lexical entry (e.g. `N0 V Adv W` or `Adv parlant, P`) and of all internal morphosyntactic structures, that is `Adv` for all simple adverbs or `Prép1 Det1 C1 Prép2 Det2 C2` for compound adverbs like *jusqu'à la fin des temps* 'until the end of time', but also `Prép1 Det1 C1` for its variant without prepositional noun phrase modifier, e.g. *jusqu'à la fin* 'until the end'.

So, we first extended *LGLex* with respect to adverbial entries by using various types of features that are encoded in the tables of both simple and compound adverbs. We added 11,328 entries (+108%), so the lexicon is now composed of 21,815 adverbial entries in total.

---
[8] http://infolingu.univ-mlv.fr/english > Language Resources > Lexicon-Grammar > Download.

## 3.1 Using the paraphrasing features

The viewpoint adverb *linguistiquement* 'linguistically' accepts the following paraphrasing constructions (Sekine, 2005) that are encoded in the table ADVMP by means of binary features:

   *linguistiquement* 'linguistically'
= *(du+d'un+au) point de vue linguistique*
   'from a linguistic point of view'
= *du point de vue de la linguistique*
   'from the point of view of linguistics'
= *au niveau linguistique*
   'at the linguistic level'
= *(au+sur le) plan linguistique*
   'on the linguistic level'
= *en linguistique*
   'in linguistics'
= *linguistiquement parlant*
   'linguistically speaking'.

For this work, we first dealt with paraphrasing constructions that are described directly in the LG tables through explicit features, as it is shown in Table 4:

| Paraphrasing features encoded in the LG tables | Lexical values of the paraphrasing features in the script |
|---|---|
| Adj-ment = avec Adj-n | "avec @Adj-n@" |
| Adj-ment = (du+d'un) point de vue Adj | "d'un point de vue @Adj@", "du point de vue @Adj@" |
| Adj-ment = du point de vue de Ddef Ndomaine | "du point de vue de Ddef @Ndomaine@" |
| Adj-ment = au niveau Adj | "au niveau @Adj@" |
| Adj-ment = au point de vue Adj | "au point de vue @Adj@" |
| Adj-ment = au plan Adj | "au plan @Adj@" |
| Adj-ment = sur le plan Adj | "sur le plan @Adj@" |
| Adj-ment = de source Adj | "de source @Adj@" |
| Adj-ment = en Adj-n | "en @Adj-n@" |
| Adj-ment = en Ndomaine | "en @Ndomaine@" |
| Adj-ment = en termes Adj | "en termes @Adj@" |
| Adj-ment = en toute Adj-n | "en toute @Adj-n@" |

Table 4. Paraphrasing features encoded directly in LG tables

The notation `@...@` specifies the lexical value of a lexical feature which is encoded systematically in a LG table. For instance, when `@Adj@` refers to table ADVMP, it can take the following values: *linguistique* 'linguistic', *politique* 'political', *informatique* 'computational', etc.

Thanks to these features, we added to *LGLex* 2,084 adverbial entries (+20%).

However, a certain number of paraphrases are part of construction features, and thus need to be extracted from them. Construction features are encoded in LG tables by means of binary features, as it is shown in Table 5:

| Construction features encoded in the LG tables | Lexical values of the paraphrasing features in the script |
|---|---|
| `Adv parlant, P` | `"@<ENT>Adv@ parlant"` |
| `N0 V W C-a-ment` | `"@C-a-ment@"` |
| `N0 V W avec Adj-n` | `"avec @Adj-n@"` |
| `N0 V W de (E+une) (façon + manière) C-a` | `"de façon @C-a@", "de manière @C-a@", "d'une façon @C-a@", "d'une manière @C-a@"` |
| `N0 V W de (E+une) (façon + manière) Adj` | `"de façon @Adj@", "de manière @Adj@", "d'une façon @Adj@", "d'une manière @Adj@"` |
| `N0hum V W avec Adj-n` | `"avec @Adj-n@"` |
| `N0hum V W de (E+une) (façon + manière) Adj` | `"de façon @Adj@", "de manière @Adj@", "d'une façon @Adj@", "d'une manière @Adj@"` |
| `à Adv parler, P` | `"à @<ENT>Adv@ parler"` |

Table 5. Paraphrasing features embedded in construction features

Using this type of features, we enhanced *LGLex* with 7,125 adverbial entries (+68%). Using both types of paraphrasing features, the number of adverbial entries in *LGLex* raised from 10,487 to 19,695 (+88%).

### 3.2 Other structures

Within the `other structures`, we distinguish three types of features: deletion, permutation and transformation. They are all described in the following sections.

#### 3.2.1 Using the deletion features

The adverb *jusqu'à la fin des* (=de les) *temps* 'until the end of time' is defined by the morpho-syntactic structure `Prép1 Det1 C1 Prép2 Det2 C2`. It accepts also the substructure *jusqu'à la fin* 'until the end', which is obtained after deletion of the prepositional noun phrase modifier *des* (=de les) *temps* 'of time', and without loss of information. The new adverb has the structure `Prép1 Det1 C1`. Table 6 displays the deletion features present in LG tables of adverbs:

| Deletion features encoded in the LG tables | Lexical value of the deletion features in script |
|---|---|
| `ConjC Det1 C1 Prép2 C2 = Det1 C1 Prép2 C2` | `"@<ENT>Det1@ @<ENT>C1@ @<ENT>Prép2@ @<ENT>C2@"` |
| `Prép Det C` | `"@<ENT>Prép@ @<ENT>Det@ @<ENT>C@"` |
| `Prép1 Det1 C1` | `"@<ENT>Prép1@ @<ENT>Det1@ @<ENT>C1@"` |

Table 6. Deletion features encoded in LG tables

The notation `@<ENT>...@` specifies the lexical value of an entry described in a LG table. In the case of compound adverbs, an entry is composed by two or more lexical components.

Thanks to the deletion features, we added to *LGLex* 1,519 adverbial entries (+14%).

#### 3.2.2 Using the permutation features

The adverb *dans un avenir proche* 'in a near future', which is defined by the morphosyntactic structure `Prép Det C Modif pré-adj Adj`, can also take the form *dans un proche avenir* due to the permutation of the adjectival phrase *proche* 'near'. The new adverb has the structure `Prép Det Modif pré-adj Adj C`, as it is shown in the Table 7:

| Permutation features encoded in the LG tables | Lexical value of the permutation features in the script |
|---|---|
| `Prép Det Modif pré-adj Adj C` | `"@<ENT>Prép@ @<ENT>Det@ @<ENT>Modif@ @<ENT>pré-adj@ @<ENT>Adj@ @<ENT>C@"` |
| `Prép Modif pré-adj Adj C` | `"@<ENT>Prép@ @<ENT>Modif@ @<ENT>pré-adj@ @<ENT>Adj@ @<ENT>C@"` |

Table 7. Permutation features encoded in LG tables

Let us consider now the compound adverb *dans les délais les plus brefs* 'as soon as possible', which is defined by the morphosyntactic structure `Prép Det C Modif pré-adj Adj`. Although it accepts the permutation of the adjectival phrase *les plus brefs* 'the shortest', it produces an agrammatical entry **dans les les plus brefs délais*. This happens only when `Modif pré-adj` corresponds to a complex unit, that is, in our example, the superlative determiner *les plus* 'the most'. To remedy this situation, we added to table PCA a specific feature enabling both the deletion and permutation of certain components without loss of information, i.e. the feature `Prép Modif pré-adj Adj C`.

This kind of problem is rather widespread since multiword units raise specific difficulties as regards their representation[9] in an electronic resource. Semantically, by definition, compound adverbs cannot be decomposed into simple units. In other words, the resulting overall meaning of the compound adverb cannot usually be deduced from the sum of the meaning of its component elements. Consequently, the various lexical components of compound adverbs are sometimes represented in the tables in an ambiguous or even arbitrary way. This is due, of course, to their irregular syntax and internal word combination constraints.

By means of the permutation features, we enhanced *LGLex* with 103 adverbial entries (+1%).

### 3.2.3 Using transformational features

Finally, `other structures` can result from general transformations applied to the free prepositional noun phrase modifier *de N* 'of N', which is part of semi-fixed adverbial expressions. For instance, the adverb *pour le bénéfice* 'for the benefit', defined by the morphosyntactic structure `Prép1 Det1 C1 de N2`, can take the following two forms: *pour le bénéfice général* 'for the general benefit' and *pour son benefice* 'for his benefit'. Table 8 displays the transformational features that are encoded in the corresponding LG tables:

| Transformation features encoded in the LG tables | Lexical value of the transformational features in the script |
|---|---|
| `Prép1 Det1 C1 de N2 = Prép1 Det1 C1 général` | `"@Prép1@ @Det1@ @C1@ général"` |
| `Prép1 Det1 C1 de N2 = Prép1 Poss2 C1` | `"@Prép1@ Poss2 @C1@"` |

Table 8. Transformational features encoded in LG tables

Thanks to these features, we added to *LGLex* 288 adverbial entries (+3%).

Taking into account the three different types of features included in `other structures`, the number of adverbial entries in *LGLex* increased from 10,487 to 12,397 (+18%).

### 3.2.4 Using the intensifying features

The focus adverb *particulièrement* 'particularly' can be modified by two specific intensifiers conveying a greater emphasis to its meaning, and thus produce the two following entries: *tout particulièrement* 'quite particularly' and *plus particulièrement* 'more particularly'. Intensifying features are represented in LG tables as shown in Table 9:

| Intensifying features encoded in the LG tables | Lexical value of the intensifying features in the script |
|---|---|
| `bien Adv` | `"bien @<ENT>Adv@"` |
| `fort Adv` | `"fort @<ENT>Adv@"` |
| `plus Adv` | `"plus @<ENT>Adv@"` |
| `tout Adv` | `"tout @<ENT>Adv@"` |
| `très Adv` | `"très @<ENT>Adv@"` |
| `(plus+moins) Adv` | `"plus @<ENT>Adv@", "moins @<ENT>Adv@"` |

Table 9. Intensifying features encoded in LG tables

Thanks to these features, the number of adverbial entries in *LGLex* augmented from 10,487 to 10,697 (+2%).

## 4 Evaluation of the extended *LGLex*

The Table 10 shows the number of the initial adverbial entries in *LGLex* and the detail of the 11,328 new entries:

---

[9] According to Gross (1986: 1), the unit of representation in a linear lexicon is roughly the word, defined as a sequence of letters separated from neighboring sequences by boundary blanks or other kind of separators, e.g. apostrophe. As a consequence, multiword units cannot be put directly into a lexicon the way simple words are. An identification procedure is needed for their occurrences in texts, and this procedure will make use of the various simple lexical components included in the multiword unit (and described in the LG tables at the elementary level of sequences of parts-of-speech). Hence, the formal linguistic properties of multiword units will determine both the procedure of identification in texts and the type of storage they require.

| Initial entries | 10,487 |
|---|---|
| Paraphrases | 9,208 |
| Other structures | 1,910 |
| Intensifying features | 210 |
| Final entries | 21,815 |

Table 10. Number of entries in *LGLex*

The results are quite satisfactory as we obtain more than double, precisely 108% new entries in the lexicon, only by exploiting precise linguistic information of high coverage, which is freely available in existing resources.

We manually evaluated the new entries of the lexicon in order to detect errors.

Indeed, a manual validation was necessary when the generated entries corresponded to one single word. For example, the adverb *pourboire compris* 'tip included', which is defined by the morphosyntactic structure `Prép Det C Modif pré-adj Adj`, accepted the deletion of the participial phrase modifier *compris* 'included', but produced a non-adverbial entry \**pourboire* 'tip'.

Then, we mention the problem of duplicates that may be produced once the generation of new adverbial entries completed. They concern a few pairs of tables, notably, the pairs PCDC and PCDN, PCA and PAC or PDETC. For example, *ces temps derniers* 'recently', defined by the morphosyntactic structure `Prép Det C Modif pré-adj Adj` and encoded in table PCA, can also take the form *ces derniers temps* due to the permutation of the adjective *derniers* 'recent'. This latter is already encoded in table PAC.

We can also evoke the case of the deleting relation that associates the adverbs of table PCDC with those of table PCDN. For example, *en l'état actuel des choses* 'in the current state of things' accepts also the substructure *en l'état actuel*[10] 'in the current state', which is obtained after deletion of the prepositional noun phrase modifier *des choses* 'of things', and without loss of information. The generated substructure is already an entry of table PCDN. In a similar manner, *en l'état actuel des connaissances* 'in the current state of knowledge' produces the same substructure. Moreover, the adverbs *dans l'état actuel des choses* 'in the current state of things' and *dans l'état actuel des connaissances* 'in the current state of knowledge', which are both encoded in

---

[10] Normally, in table PCDN, the fixed part of the semi-fixed adverbial expression comprises also the second preposition *de* 'of' which is constant for all entries of the table, and for that reason it is not encoded explicitly.

table PCDC, accept the substructure *dans l'état actuel* 'in the current state'.

In fact, each substructure provides information about the corresponding entry, and the generated substructures can be filtered automatically to easily delete duplicates.

Last, errors in the new entries are sometimes due to the way initial adverbial entries are encoded in tables. Considering the adverb *à cette heure-ci* 'at the present time' represented in table PCA: the noun component *heure* 'time' is encoded together with the hyphen, and thus form an amalgam that is automatically reproduced in the substructure *à cette heure-* 'at this time'.

## 5 Conclusion and future work

At a time when the lack of large scale lexical syntactic resources for French impedes on NLP research, we showed the interest of using fine-grained linguistic information, which is provided in existing resources, in order to enrich or diversify their content. This work led to an increase of 108% of the adverbial entries in *LGLex*.

These encouraging results confirm it is worthwhile exploiting features such as paraphrases. Therefore, we plan to complete the LG tables in that direction, starting, for example, with the table of verbal manner adverbs:

> Adj-ment = en tout Nabstrait =:
> amicalement = en toute amitié
> 'friendly'          'in all friendship'
>
> Adj-ment = par Nmoyen_communication =:
> téléphoniquement = par téléphone
> 'by telephone'      'by telephone'

Furthermore, we plan to convert the new adverbial entries into the Le*fff* format (Sagot, 2010), in order to integrate them into a parser, following similar work by Tolone and Sagot (2011) and Tolone (2011).

Besides, the new adverbial entries will be added to the French morphological lexicon DELA, which will enable us to evaluate their accuracy by means of both a corpus' annotation practice and a detailed comparison with related work by Laporte *et al.* (2008).

Finally, we can also consider enhancing the French Wordnet with respect to adverbial entries (Sagot *et al.*, 2009).